\documentclass[a4paper]{article}
\usepackage[utf8]{inputenc}
\usepackage{INTERSPEECH2018}
\usepackage{cite}
\usepackage{times}
\usepackage{latexsym}
\usepackage{color}
\usepackage{listings}
\usepackage{url}
\usepackage{hyperref}
\usepackage{cleveref}
\usepackage{textcomp}
\usepackage{caption}
\usepackage{mathtools}
\usepackage{wrapfig}
\usepackage{multirow}
\usepackage{setspace}

\DeclareUnicodeCharacter{00A0}{}

\newcommand{\crnn}{RETURNN}

\Crefname{equation}{Eq.}{Eqs.}
\Crefname{figure}{Fig.}{Figs.}
\Crefname{tabular}{Tab.}{Tabs.}

\makeatletter

\renewcommand{\section}{\@startsection
  {section}%
  {1}%
  {}%
  {-0.7\baselineskip}%
  {0.3\baselineskip}%
  {}}%

\renewcommand{\subsection}{\@startsection
  {subsection}%
  {2}%
  {}%
  {-0.7\baselineskip}%
  {0.3\baselineskip}%
  {}}%

\renewcommand{\subsubsection}{\@startsection
  {subsubsection}%
  {3}%
  {}%
  {-0.7\baselineskip}%
  {0.3\baselineskip}%
  {}}%

\g@addto@macro\normalsize{%
  \setlength\abovedisplayskip{4pt plus 2pt minus 2pt}
  \setlength\belowdisplayskip{4pt plus 2pt minus 2pt}
  \setlength\abovedisplayshortskip{4pt plus 2pt minus 2pt}
  \setlength\belowdisplayshortskip{4pt plus 2pt minus 2pt}
}

\makeatother

\title{Improved training of end-to-end attention models for speech recognition}
\name{Albert Zeyer$^{1,2,3}$, Kazuki Irie$^1$, Ralf Schlüter$^1$, Hermann Ney$^{1,2}$}
\address{
$^1$Human Language Technology and Pattern Recognition,
Computer Science Department, \\
RWTH Aachen University, 52062 Aachen, Germany, \\
$^2$AppTek, USA, \url{http://www.apptek.com/}, \\
$^3$NNAISENSE, Switzerland, \url{https://nnaisense.com/}}
\email{\{zeyer, irie, schlueter, ney\}@cs.rwth-aachen.de}

\begin{document}

\maketitle
\begin{abstract}
Sequence-to-sequence attention-based models on subword units
allow simple open-vocabulary end-to-end speech recognition.
In this work, we show that such models can achieve competitive
results on the Switchboard 300h and LibriSpeech 1000h tasks.
In particular, we report the state-of-the-art word error rates (WER)
of 3.54\% on the dev-clean and 3.82\% on the test-clean evaluation subsets of LibriSpeech.
We introduce a new pretraining scheme by starting with
a high time reduction factor and lowering it during training,
which is crucial both for convergence and final performance.
In some experiments, we also use an auxiliary CTC loss function
to help the convergence. In addition, we train long short-term
memory (LSTM) language models on subword units.
By shallow fusion, we report up to 27\% relative improvements in WER
over the attention baseline without a language model.
\end{abstract}
\noindent\textbf{Index Terms}: attention, end-to-end, speech recognition

\section{Introduction}

Conventional speech recognition systems \cite{zeyer17:lstm}
with neural network (NN) based acoustic models
using the hybrid hidden Markov models (HMM) / NN approach
\cite{bourlard1994hybrid,robinson1994rnnhmm}
usually operate on the phone level,
given a phonetic pronunciation lexicon (from phones to words).
They require
a pretraining scheme with HMM
and Gaussian mixture models (GMM) as emission probabilities
to bootstrap good alignments of the HMM states.
Context-independent phones are used initially
because context-dependent phones need a good clustering,
which is usually created on good existing alignments
(via a Classification And Regression Tree (CART) clustering \cite{young1994cart}).
This boot-strapping process is iterated a few times.
Then a hybrid HMM / NN is trained with
frame-wise cross entropy.
Recognition with such a model requires a sophisticated beam search decoder.
Handling out-of-vocabulary words is also not straightforward and increases the complexity.
There was certain work to remove the GMM dependency in the pretraining
\cite{senior2014gmmfree},
or to be able to train without an existing alignment
\cite{zeyer2017:ctc,povey2016lfmmi,sak2015ctc},
or to avoid the lexicon \cite{kanthak:icassp2002},
which simplifies the pretraining procedure
but still is not end-to-end.

An \emph{end-to-end model} in speech recognition generally denotes a simple single model
which can be trained from scratch,
and usually directly operates on words, sub-words or characters/graphemes.
This removes the need for a pronunciation lexicon
and the whole explicit modeling of phones,
and it greatly simplifies the decoding.

Connectionist temporal classification (CTC) \cite{graves2006ctc}
has been often used as an end-to-end model for speech recognition,
often on characters/graphemes
\cite{graves14endtoendspeech,hannun2014deepspeech,miao2015eesen,
collobert2016wav2letter,zhou2017policylearning,zweig2017allneural}
or on sub-words \cite{liu2017gramctc}
but also directly on words \cite{soltau2017wordctc,audhkhasi2017wordctc}.

The \emph{encoder-decoder framework with attention} has become the standard
approach for machine translation \cite{bahdanau2014nmt,luong2015attentionmt,wu2016google}
and many other domains such as images \cite{ba2014dram}.
Recent investigations have shown promising results by applying the same approach for
speech recognition \cite{
chan2016las, 
doetsch2016:bidir-dec-att, 
chiu2017sotaasratt, 
battenbergAsru17,
toshniwal2017mltatt
}.
In this work, we also investigate techniques to improve
recurrent encoder-attention-decoder based systems for speech recognition.
We use long short-term memory (LSTM) neural networks \cite{hochreiter1997lstm}
for the encoder and the decoder.
Our model is similar to the architecture used in machine translation \cite{zeyer2018:returnn},
except of encoder time reduction.
This generality of the model and the simplicity is its strength.
Although a valid argument against this model for speech recognition
is that it is in fact too powerful because it does not require monotonicity
in its implicit alignments.
There are attempts to restrict the attention to become monotonic in various ways
\cite{chorowski2014end2end,jaitly2016onlineseq2seq,
aharoni2016hardmonatt,
raffel2017onlinemonotonic,chiu2017monotonic,tjandra2017localmonotonicatt,
prabhavalkar2017attanalysis, 
hou2017gaussian
}.
In this work, our models are without these modifications and extensions.

Recently, alternative models for end-to-end modeling were also suggested,
such as
inverted HMMs \cite{DoetschHannemannSchlueter17:InvHMM},
the recurrent transducer \cite{
rao2017transducerasr,
battenberg2017neuraltransducerasr,
prabhavalkar2017compseq2seq},
or the recurrent neural aligner \cite{sak2017neuralaligner}.
In many ways, these can all be interpreted in the same encoder-decoder-attention framework,
but these approaches often use some variant of hard latent monotonic attention
instead of soft attention.

Our models operate on \emph{subword units} which are created
via \emph{byte-pair encoding (BPE)} \cite{sennrich16bpe}.
We introduce a \emph{pretraining scheme} applied on the encoder,
which grows the encoder in layer depth,
as well as decreases the initial high encoder time reduction factor.
To the best of our knowledge, we are the first to apply pretraining
for encoder-attention-decoder models.
We use \crnn{} \cite{zeyer2018:returnn,doetsch2017:returnn}
based on TensorFlow \cite{tensorflow2015} for its computation.
We have implemented our own flexible and efficient beam search decoder
and efficient LSTM kernels in native CUDA.
In addition, we train subword-level LSTM language models \cite{sundermeyer2012lstm}, which
we integrate in the beam search by \emph{shallow fusion} \cite{gulcehre2015using}.
The source code is fully open\footnote{\tiny\url{https://github.com/rwth-i6/returnn}},
as well as all the setups of the experiments in this paper%
\footnote{\tiny\url{https://github.com/rwth-i6/returnn-experiments/tree/master/2018-asr-attention}}.
We report competitive results on the 300h-Switchboard
and LibriSpeech \cite{panayotov2015librispeech}.
In particular on Librispeech,
our system achieves WERs of 3.54\% on the dev-clean and 3.82\% on the test-clean evaluation subsets,
which are the best results obtained on this task to the best of our knowledge.

\section{Pretraining}
Compared to machine translation, the input sequences are much longer in speech recognition,
relatively to the output sequence
(e.g.\ with BPE 10K subword units, and audio feature frames every 10ms,
more than 30 times longer on Switchboard on average).
However, as the original input is continuous,
some sort of downscaling in the time dimension works,
such as
concatenation in the feature dimension of consecutive time-frames
\cite{chan2016las,prabhavalkar2017compseq2seq,povey2016lfmmi,pundak2016lfr}.
We use max-pooling in the time-dimension which is simpler.
The time reduction can be done directly on the features
or alternatively at multiple steps inside the encoder,
e.g.\ after every encoder layer \cite{chan2016las}.
This is also what we do.
This allows the encoder to better compress any necessary information.

We observed that a high time reduction factor makes the training much simpler.
In fact, without careful tuning, usually the model will not converge
without a high time reduction factor (16 or 32),
as it was also observed in the literature \cite{chan2016las}.
However, we also observed that a low time reduction factor (e.g. 8)
can perform better after all, when pretrained with a high time reduction factor.

Also, it has been shown that deep LSTM models can benefit from layer-wise pretraining,
by starting with 1 or 2 layers and adding more and more layers \cite{zeyer17:lstm}.
We apply the same pretraining.

To improve the convergence further, we disable label smoothing during pretraining
and only enable it after pretraining.
Also, we disable dropout during the first few pretraining epochs in the encoder.

\section{Model}

We use a deep bidirectional LSTM encoder network,
and LSTM decoder network.
After every layer in the encoder, we optionally do max-pooling in the time dimension
to reduce the encoder length.
I.e.\ for the input sequence $x_1^T$,
we end up with the encoder state
\[
h_1^{T'} = \operatorname{LSTM}_{\#\rm{enc}}
\circ \cdots
\circ \operatorname{max-pool}_1
\circ \operatorname{LSTM}_1 (x_1^T) ,
\]
where $T' = \rm{red} \cdot T$
for the time reduction factor $\rm{red}$,
and $\#\rm{enc}$ is the number of encoder layers,
with $\#\rm{enc} \ge 2$.
We use the MLP attention \cite{bahdanau2014nmt,chorowski2014end2end,
luong2015attentionmt,
jaitly2016onlineseq2seq,bahar17iwslt}.
Our model closely follows the machine translation model
presented by Bahar et al.\ \cite{bahar17iwslt}
and Bahdanau et al.\ \cite{bahdanau2014nmt}
and we use a variant of attention weight / fertility feedback \cite{tu2016ACL},
which is inverse in our case, to use a multiplication instead of a division,
for better numerical stability.
More specifically, the attention energies $e_{i,t} \in \mathbb{R}$
for encoder time-step $t$
and decoder step $i$ are defined as
\[
e_{i,t} = v^\top \tanh (W [s_i, h_t, \beta_{i,t}] ) ,
\]
where $v$ is a trainable vector, $W$ a trainable matrix,
$s_i$ the current decoder state, $h_t$ the encoder state,
and $\beta_{i,t}$ is the attention weight feedback,
defined as
\[
\beta_{i,t} = \sigma(v_\beta^\top h_t ) \cdot \sum_{k=1}^{i-1} \alpha_{k,t},
\]
where $v_\beta$ is a trainable vector.
Then the attention weights are defined as
\[ \alpha_{i} = \operatorname{softmax}_t (e_i) \]
and the attention context vector is given as
\[ c_i = \sum_t \alpha_{i,t} h_t . \]
The decoder state is recurrent function implemented as
\[ s_i = \operatorname{LSTMCell} (s_{i-1}, y_{i-1}, c_{i-1}) \]
and the final prediction probability for the output symbol $y_i$ is given as
\[ p(y_i | y_{i-1}, x_1^T) =
\operatorname{softmax} ( \operatorname{MLP}_{\rm{readout}}
(s_i, y_{i-1}, c_i) ) . \]
In our case we use $\operatorname{MLP}_{\rm{readout}}
= \operatorname{linear} \circ \operatorname{maxout} \circ \operatorname{linear}$.

\section{Sub-word units}
Characters/graphemes are probably the most generic and simple output units for generating texts
but it has been shown that sub-word units can perform better \cite{chiu2017sotaasratt}
and they can be just as generic since the characters can be included in the set of subword units.
Using words as output units is also possible but it does not allow to recognize out-of-vocabulary words
and it requires a large softmax output and thus is computational expensive.
An inhomogeneous length distribution
as well as an imbalance in the label occurence can also make training harder.

In all the experiments, we use byte-pair encoding (BPE)
\cite{sennrich16bpe}
to create subword units,
which are the output targets of the decoder.
The beam search decoding will go over these BPE units,
and then select the best hypothesis. Therefore, our system is open-vocabulary.
At the end of decoding, the BPE units are merged into words in
order to obtain the best hypothesis on word level.
In addition, we add the special tokens from the transcriptions which denote noise, vocalized-noise
and laughter in our BPE vocabulary set. Our recognizer can also potentially recognize these special events.

\section{Language model combination}
We also improve the recognition accuracy of our recognizer using external language models.
We train LSTM language models \cite{sundermeyer2012lstm} on the same BPE vocabulary set
as the end-to-end model, using \crnn{} with TensorFlow.
For Switchboard, the training set of 27M words
concatenating Switchboard and Fisher parts of transcriptions was used.
For LibriSpeech, we use the 800M-word dataset officially available%
\footnote{\tiny\url{http://www.openslr.org/11/}}
for training language models.
It can be noted that in the case of Switchboard, there is some overlap between
the training data for language models and the transcription used to train the end-to-end model:
3M out of 27M words are used to train the end-to-end system.
While for the LispriSpeech, 800M-word data is fully external
to the end-to-end models.
Our experiments show that this difference in amount of external data
directly affects the performance
improvements by the use of external language model.
For both tasks, we use a LSTM LM with one input projection layer size of 512 dimension and two
LSTM layers with 2048 nodes.
We apply dropout at the input of all hidden layers with the rate of 0.2.
The standard stochastic gradient descent with global gradient clipping is used
for optimization to train all LSTM LMs.

We integrate the external language model in the beam search by shallow fusion
\cite{gulcehre2015using}.
The weight for the language model has been optimized
by grid search on the development set WER.
We found 0.23 and 0.36 to be optimal respectively
for Switchboard and LibriSpeech
(the weight on the attention model is 1).

For LibriSpeech, we also train Kneser-Ney smoothed $n$-gram count based language models \cite{kneser1995improved} on
the same BPE vocabulary set using SRILM toolkit \cite{sri:02}.
The comparison of perplexities can be found in Table \ref{tab:lib_ppl}.
We also report WERs using the 4-gram count model by shallow fusion with a weight of 0.01, for comparison
to the performance of LSTM LM.


\begin{table}[th]
	\setlength{\tabcolsep}{0.3em}
	\caption{Perplexities (PPL) on the concatenation of dev-clean and dev-other sets of LibriSpeech. All models
		have the same vocabulary of 10K BPE.
	}
	\label{tab:lib_ppl}
	\centerline{
		\begin{tabular}{|c|r|r|r|r|}
			\hline
			LM & 3-gram & 4-gram & 5-gram & LSTM \\ \hline
			PPL & 104.6 & 88.2 & 85.1 & 65.9 \\ \hline
		\end{tabular}}
	\end{table}


\section{Experiments}

All attention models and neural network language models
were trained and decoded with \crnn{}.
For both Switchboard and LibriSpeech, we first used the BPE vocabulary of 10K subword
units to tune the hyperparameters of the model, then trained the models with 1K and 5K
BPE units.
We found 1K and 10K to be optimal for Switchboard and LibriSpeech respectively.
We use label smoothing \cite{szegedy2016labelsmth},
dropout \cite{srivastava2014dropout},
Adam \cite{kingma2014adam},
learning rate warmup \cite{chiu2017sotaasratt},
and automatic learning rate scheduling according to a cross-validation set
("Newbob") \cite{zeyer17:lstm}.

\subsection{Pretraining}

In all cases we use layer-wise pretraining for the encoder,
where we start with two encoder layers and a single max-pool in between with factor 32.
Then we add a LSTM layer and a max-pool in between,
and we reduce the first max-pool to factor 16 and the new one with factor 2
such that we always keep the same total encoder time reduction factor of 32.
Only when we end up at 6 layers, we remove some of the max-pooling ops to
get a final total time reduction factor of e.g.\ 8.
Directly starting with a time reduction factor of 8 with and with 2 layers did not work for us.
Also directly starting with 6 layers and time reduction factor of 32 did not work for us.
Similar experiments for translation converged also without pretraining,
however with much worse performance compared when layer-wise pretraining was used
\cite{zeyer2018:returnn}.
With more careful tuning or more training data, it might have worked without pretraining
as it is seen in the literature,
however, that is not necessary with pretraining.

We were interested in the optimal final total time reduction factor,
after the pretraining with time reduction factor 32.
We tried factor 8, 16 and 32,
and ended up with 20.4, 21.0 and 21.9 WER\% respectively,
on the full Hub5'00 set (Switchboard + Callhome).
Thus we continue to use a final reduction factor of 8 in all further experiments.
Note that a lower factor requires more memory and more computation
for the global attention and was not feasible
with our hardware and computational resources.

\subsection{Switchboard 300h}

Switchboard consists of about 300 hours of training data.
There is also the additional Fisher training dataset, so combined it makes the total of about 2000h.
In this work, we only use the 300h-Switchboard training data.
We use 40-dimensional Gammatone features \cite{Schlueter2007},
and the feature extraction was done with RASR \cite{wiesler2014:rasr}.
Results are shown in \Cref{tab:swb}.
We observe that our attention model performs better
on the easier Switchboard subset of the dev set Hub5'00,
where it is the best end-to-end model we know.
On the harder Callhome part, it also performs well compared to other end-to-end models
but the relative difference is not as high.

\begin{table}[th]
\setlength{\tabcolsep}{0.3em}
\caption{Comparisons on Switchboard 300h.
The hybrid HMM/NN model is a 6 layer deep bidirectional LSTM.
The attention model has a 6 layer deep bidirectional LSTM encoder
and a 1 layer LSTM decoder.
CDp are (clustered) context-dependend phones.
Byte-pair encoding (BPE) are sub-word units.
SWB and CH are from Hub5'00.
$^1$added noise from external data.
$^2$added the lexicon, i.e.\ also additional data.
}
\label{tab:swb}
\centerline{
	\resizebox{\columnwidth}{!}{
\begin{tabular}{|c|c|c|c|c|c|}
\hline
\multirow{2}{*}{model} & \multirow{2}{*}{LM} & label  & \multicolumn{3}{c|}{WER[\%]} \\
&& unit & SWB & CH & Hub5'01 \\ \hline
\hline
\href{http://www.danielpovey.com/files/2016_interspeech_mmi.pdf}{LF MMI, 2016} \cite{povey2016lfmmi} & 4-gram & CDp & 9.6 & 19.3 & \\ \hline
hybrid & 4-gram & CDp & 9.8 & 19.0 & 14.7 \\ \hline
hybrid & LSTM & CDp & \textbf{8.3} & \textbf{17.3} & 12.9 \\ \hline
\hline
\href{https://arxiv.org/abs/1412.5567}{CTC$^1$, 2014} \cite{hannun2014deepspeech} & RNN & chars & 20.0 & 31.8 & \\ \hline
\href{http://ai.stanford.edu/~amaas/papers/ctc_clm_naacl_2015.pdf}{CTC, 2015} \cite{maas2015lexicon} & none & chars & 38.0 & 56.1 & \\ \hline
\href{http://ai.stanford.edu/~amaas/papers/ctc_clm_naacl_2015.pdf}{CTC, 2015} \cite{maas2015lexicon} & RNN & chars & 21.4 & 40.2 & \\ \hline
\href{http://homepages.inf.ed.ac.uk/llu/pdf/llu_icassp16.pdf}{attention, 2016} \cite{lu2016att} & none & chars & 32.8 & 52.7 & \\ \hline
\href{http://homepages.inf.ed.ac.uk/llu/pdf/llu_icassp16.pdf}{attention, 2016} \cite{lu2016att} & 5-gram & chars & 30.5 & 50.4 & \\ \hline
\href{http://homepages.inf.ed.ac.uk/llu/pdf/llu_icassp16.pdf}{attention, 2016} \cite{lu2016att} & none & words & 26.8 & 48.2 & \\ \hline
\href{http://homepages.inf.ed.ac.uk/llu/pdf/llu_icassp16.pdf}{attention, 2016} \cite{lu2016att} & 3-gram & words & 25.8 & 46.0 & \\ \hline
\href{https://arxiv.org/abs/1609.05935}{CTC, 2017} \cite{zweig2017allneural} & none & chars & 24.7 & 37.1 & \\ \hline
\href{https://arxiv.org/abs/1609.05935}{CTC, 2017} \cite{zweig2017allneural} & $n$-gram & chars & 19.8 & 32.1 & \\ \hline
\href{https://arxiv.org/abs/1609.05935}{CTC$^2$, 2017} \cite{zweig2017allneural} & word RNN & chars & 14.0 & \textbf{25.3} & \\ \hline
\href{http://www.isca-speech.org/archive/Interspeech_2017/pdfs/1118.PDF}{attention, 2017} \cite{toshniwal2017mltatt} & none & chars & 23.1 & 40.8 & \\ \hline
\hline
%
\multirow{3}{*}{attention} & \multirow{2}{*}{none} & BPE 10K & 13.5 & 27.1 & 19.9 \\ \cline{3-6}
 &  & BPE 1K & 13.1 & 26.1 & 19.7 \\ \cline{2-6}
 & LSTM & BPE 1K & \textbf{11.8} & 25.7 & 18.1 \\ \hline
\end{tabular}}}
\end{table}

\subsection{LibriSpeech 1000h}

LibriSpeech training dataset consist of about 1000 hours of read audio books.
The dev and test sets were split into simple ("clean")
and harder ("other") subsets \cite{panayotov2015librispeech}.
We do 40-dim.\ MFCC feature extraction
on-the-fly in \crnn{}, based on librosa \cite{librosa_mcfee_brian_2017_293021}.
We use CTC as an additional loss function applied
on top of the decoder to help the convergence,
although this is not used in decoding \cite{hori2017attctc}.
We initially trained only using the train-clean set
and restricting it to sequences not longer than 75 characters in the orthography.
Results are shown in \Cref{tab:librispeech}.
Our end-to-end system achieves competitive performance even without using language models.
We observed that the shallow fusion with LSTM LM brings from 17\% to 27\% relative
improvements in terms of WER on different subsets. This improvement is much larger
than in the case of Switchboard. The amount of data is most likely the reason
for this observation. For Librispeech, the external data of 800M words is used
to train the language models, which is 80 times larger than the 10M words corresponding to
the transcription of 1000 hours of audio. In addition, this 10M transcription is not part of
the language model training data. In case of Switchboard, the LM is trained only on about 27M words,
including 3M of transcription used to train the end-to-end system.
Text data for conversational speech is not as readily available as for read speech.
The WER of 3.54\% on the dev-clean and 3.82\% on the test-clean subsets are the best performance on this task
to the best of our knowledge for systems trained only using LibriSpeech data.

\begin{table}[th]
\setlength{\tabcolsep}{0.3em}
\caption{Comparisons on LibriSpeech 1000h.
The attention model has a 6 layer deep bidirectional LSTM encoder
and a 1 layer LSTM decoder.
CDp are (clustered) context-dependend phones.
Byte-pair encoding (BPE) are sub-word units.
Lattice-free (LF) maximum mutual information (MMI) \cite{povey2016lfmmi}
is a sequence criterion to train a hybrid HMM/NN model.
Auto SeGmentation (ASG) \cite{liptchinsky2017asg}
can be seen as a variant of the CTC criterion and model.
Policy learning is a sequence training method,
applied here on a CTC model \cite{zhou2017policylearning}.
If not specified, the official 4-gram word LM is used.
The remaining attention models are all our models.
}
\label{tab:librispeech}
\centerline{
	\resizebox{\columnwidth}{!}{
\begin{tabular}{|c|c|c|c|c|c|c|c|}
\hline
\multirow{3}{*}{model} & \multirow{3}{*}{LM} & \multirow{2}{*}{label} & \multicolumn{4}{c|}{WER[\%]} \\ \cline{4-7}
&& \multirow{2}{*}{unit} & \multicolumn{2}{c|}{dev} & \multicolumn{2}{c|}{test} \\  \cline{4-7}
&&  & clean & other & clean & other \\
\hline\hline
\href{http://www.danielpovey.com/files/2015_icassp_librispeech.pdf}{hybrid, FFNN, 2015} \cite{panayotov2015librispeech} & 4-gram & CDp & 4.90 & 12.98 & 5.51 & 13.97 \\ \hline
\href{http://www.danielpovey.com/files/2016_interspeech_mmi.pdf}{LF MMI, LSTM, 2016} \cite{povey2016lfmmi} & 4-gram & CDp & & & 4.28 & \\ \hline
\hline
\href{https://arxiv.org/abs/1512.02595}{CTC, 2015} \cite{amodei2015deepspeech2} & 4-gram & chars & & & 5.33& 13.25 \\ \hline
\href{https://arxiv.org/abs/1712.09444}{ASG (CTC), 2017} \cite{liptchinsky2017asg} & 4-gram & chars & & & 4.80 & 14.50 \\ \hline
\href{https://arxiv.org/abs/1712.09444}{ASG (CTC), 2017} \cite{liptchinsky2017asg} & none & chars & & & 6.70 & 20.80 \\ \hline
\href{https://arxiv.org/abs/1712.07101}{CTC, PL, 2017} \cite{zhou2017policylearning} & 4-gram & chars & 5.10 & 14.26 & 5.42 & 14.70 \\
 \hline
\hline
\multirow{3}{*}{attention} & none & BPE & 4.87 & 14.37 & 4.87 & 15.39 \\ \cline{2-7}
 & 4-gram & BPE & 4.79 & 14.31 & 4.82 & 15.30 \\ \cline{2-7}
 & LSTM & BPE & \textbf{3.54} & \textbf{11.52} & \textbf{3.82} & \textbf{12.76} \\ \hline  
\end{tabular}}}
\end{table}

\subsection{Beam search prune error analysis}

Beam search is an approximation for the decision rule
\[ x_1^T \rightarrow \hat{w}_1^N := \arg\max_{w_1^N} p(w_1^N|x_1^T) . \]
The approximation is the pruning we apply due to the beam size.
Beam search decoding for hybrid models
is very sophisticated and uses a dynamic beam size
based on the partial hypothesis scores
which can become very large (on the order of thousands) \cite{nolden2017:phd}.
The beam search for attention models works directly on the labels,
i.e.\ on the BPE units in our case,
and usually a static fixed very low beam size (e.g. 10) is used.
It has been shown that increasing the beam size much more
does not help in increasing the overal performance.
This indicates that we do not have a search problem
but we wanted to analyze this in more detail.
Specifically,
we are interested in how much errors we are making due to the pruning
for our attention models,
and we can count that by calculating the search score of the real target sequence,
and compare it to the search score of the decoded sequence.
If the decoded sequence has a higher score than the real target sequence,
we have not made a search error but it is a model error.
We count the number of sequences where the decoded sequence has a lower score
than the real target sequence.
We report our results in \Cref{tab:searchanalysis}.
We observe that for our standard beam size 12,
the number of search errors are well below 1\%,
and also the WER will not noticeably improve with a larger beam size.
Note that we only analyzed the search errors regarding reaching the real target sequence.
We did not count search errors regarding reaching any sequence with lower WER.
However, our results still suggest that we do not seem to have a search problem
but a model problem.

\begin{table}[th]
\setlength{\tabcolsep}{0.3em}
\caption{Beam search error analysis,
performed on LibriSpeech,
without language model.
We provide both the number of reference-related search errors,
relative to the number of sequences,
and also the corresponding WER.
}
\label{tab:searchanalysis}
\centerline{
\begin{tabular}{|c|c|c|c|c|}
\hline
\multirow{2}{*}{beam} &  \multicolumn{4}{c|}{search errors [\%] (WER [\%])} \\
\multirow{2}{*}{size} & \multicolumn{2}{c|}{dev} & \multicolumn{2}{c|}{test} \\
&  clean & other & clean & other \\
\hline\hline
4 & 1.52 (4.87) & 1.68 (14.53) & 1.07 (4.87) & 1.70 (15.49) \\ \hline
8 & 0.96 (4.88) & 0.98 (14.40) & 0.76 (4.87) & 1.02 (15.39) \\ \hline
12 & 0.81 (4.87) & 0.59 (14.37) & 0.61 (4.86) & 0.71 (15.39) \\ \hline
16 & 0.70 (4.87) & 0.52 (14.36) & 0.50 (4.86) & 0.58 (15.37) \\ \hline
32 & 0.26 (4.87) & 0.14 (14.34) & 0.19 (4.86) & 0.20 (15.34) \\
\hline
\end{tabular}}
\end{table}

\section{Conclusions}
We presentented an encoder-decoder-attention model for speech recognition operating on BPE subword units.
We introduced a new method for pretraining the encoder, which was crucial for both convergence and
the performance in terms of WER. We further improved our recognition accuracy by a joint beam search with a LSTM LM trained on the same subword vocabulary. We carried out experiments on two standard datasets. On the 300h-Switchboard,
we achieved competitve results compared to the previously reported end-to-end models, while the WERs are still
higher than the conventional hybrid systems. On the 1000h-LibriSpeech task, we obtained competitive results across different
evaluation subsets. To the best of our knowledge, the WERs of 3.54\% on the dev-clean and 3.82\% on the test-clean subsets are the best results reported on this task, when only the official LibriSpeech training data is used.

\section{Acknowledgements}
\begingroup
\begin{footnotesize}
\begin{spacing}{0.5}
\setlength{\columnsep}{5pt}%
\setlength{\intextsep}{0pt}%
This work has received funding from the European Research Council (ERC)
under the European Union's Horizon 2020 research and innovation programme
(grant agreement No 694537, project "SEQCLAS").
The work reflects only the authors' views and
the ERC Executive Agency is not responsible for any
use that may be made of the information it contains.
The GPU cluster used for the experiments was partially funded by Deutsche Forschungsgemeinschaft
(DFG) Grant INST 222/1168-1.
\end{spacing}
\end{footnotesize}
\endgroup

\bibliographystyle{IEEEtran}


\let\normalsize\footnotesize\normalsize

\let\OLDthebibliography\thebibliography
\renewcommand\thebibliography[1]{
  \OLDthebibliography{#1}
  \setlength{\parskip}{0pt}
  \setlength{\itemsep}{0pt plus 0.07ex}
}

\bibliography{mybib}

\end{document}